\documentclass[10pt,letterpaper]{article}
\usepackage[top=0.85in,left=5.cm,footskip=0.75in,marginparwidth=2in]{geometry}

\usepackage[utf8]{inputenc}

 \usepackage{natbib}
\usepackage{nameref,hyperref}

\usepackage[right]{lineno}

\usepackage{microtype}
\DisableLigatures[f]{encoding = *, family = * }

\raggedright
\usepackage{setspace} 
\usepackage{changepage}

\usepackage[aboveskip=1pt,labelfont=bf,labelsep=period,singlelinecheck=off]{caption}

\makeatletter
\renewcommand{\@biblabel}[1]{\quad#1.}
\makeatother

\usepackage{lastpage,fancyhdr,graphicx}
\usepackage{epstopdf}
\fancyheadoffset[L]{1.25in}
\fancyfootoffset[L]{1.25in}

\usepackage{color}

\definecolor{Gray}{gray}{.25}

\usepackage{graphicx}

\usepackage{sidecap}

\usepackage{wrapfig}
\usepackage[pscoord]{eso-pic}
\usepackage[fulladjust]{marginnote}
\reversemarginpar

\begin{document}


\vspace*{0.35in}

\begin{flushleft}
{\LARGE

\textbf\newline{{Error-related Potential driven Reinforcement Learning for adaptive Brain-Computer Interfaces}
}
}
 \newline
\\
Aline Xavier Fidêncio\textsuperscript{1,2,3,*},
Felix Grün\textsuperscript{1,4},
Christian Klaes\textsuperscript{3} and
Ioannis Iossifidis\textsuperscript{1,*}
\\
\bigskip
\bf{1} Robotics and BCI Laboratory, Institute of Computer Science, Ruhr West University of Applied Sciences, Mülheim an der Ruhr, Germany 
\\
\bf{2} Faculty of Electrical Engineering and Information Technology, Ruhr University Bochum, Bochum, Germany
\\
\bf{3} KlaesLab, Department of Neurosurgery, University Hospital Knappschaftskrankenhaus, Ruhr University Bochum, Bochum, Germany
\\
\bf{4} Faculty of Computer Science, Ruhr University Bochum, Bochum, Germany
\\
\bigskip
* aline.xavierfidencio@rub.de, ioannis.iossifidis@hs-ruhrwest.de

\end{flushleft}




\section*{Abstract}
    Brain-computer interfaces (BCIs) are developed for individuals with motor disabilities to offer alternative methods of communication, such as controlling and interacting with external devices.
    These systems are designed to improve quality of life, particularly for those with limited mobility. BCIs that rely on non-invasive recordings, such as electroencephalography (EEG), have been developed for a wide range of scenarios due to their practicality and safety. However, BCI performance is often limited by non-stationarities in the EEG data, which can arise from changes in the subject's mental state or device characteristics, such as electrode impedance. These challenges motivate ongoing research into the development of adaptive BCIs capable of handling such variations.
    Over the past years, the interest in using the so-called error-related potentials (ErrPs) to improve BCI performance has increased. This is because ErrPs represent the neural response to both self-made and external errors and can be measured using non-invasive techniques. These signals have been combined with different BCI paradigms and used in different works to improve BCI performance via error correction or adaptation.
    In this work we introduce and explore a new approach for creating an adaptive ErrP-based BCI by evaluating the use of reinforcement learning (RL).
    This study demonstrates the feasibility of a RL-driven adaptive brain-computer interface (BCI) framework that integrates error-related potentials (ErrPs) and motor imagery. By employing two RL agents to dynamically adapt to EEG non-stationarities, we validate the framework on both a publicly available motor imagery dataset and using a novel experimental protocol involving a fast-paced game designed to enhance user engagement. Results highlight the framework’s potential: RL agents successfully learned control policies from user interactions, achieving robust performance across datasets. However, a critical finding emerged from the game-based protocol: the use of motor imagery in a high-speed interaction paradigm proved ineffective for most participants, underscoring limitations in task design for real-time BCI applications. These outcomes emphasize the promise of RL for adaptive BCIs while identifying practical constraints in task complexity and user responsiveness.

    \bigskip
    {\bf Keywords:} error-related potentials (ErrPs), adaptive brain-computer interface, BCI, reinforcement learning (RL), motor imagery (MI), EEG

\section{Introduction}
    Rehabilitation and assistance systems can be used to improve life-quality for patients living with motor impairments caused, for example, by an amputation, spinal cord injury, or stroke~\citep{abiri_comprehensive_2019, soekadar_brainmachine_2015,kumar_review_2019}. Advances in the field of brain-computer interfaces (BCIs) provide patients with an alternative communication path to these systems. This is achieved through the direct decoding or classification of specific brain signals and their translation into appropriate control commands for the external systems. While different technologies can be used for neural signal acquisition, non-invasive electroencephalography (EEG) devices are widely applied due to their good temporal resolution, attractive price, and usability~\citep{kumar_review_2019}. 
    
    BCIs can be developed based on different experimental paradigms and used to control different devices: from a cursor on the monitor, to robotic arms, or wheelchairs, for example~\citep{kumar_review_2019}. 
    The experimental paradigms define, among others, which kind of brain signal should be decoded and common applications are based on event-related synchronization/desynchronization (ERS/ERD) modulations generated during motor imagination, steady-state visual evoked potentials (SSVEPs), or P300 potentials. For a comprehensive review on different paradigms we refer the reader to~\citep{abiri_comprehensive_2019}. 
    
    However, the classification of brain signals is a challenging task and a current limitation in developing high performance BCI systems for the long-term use is their decreasing performance over time due to the inherent non-stationarities in EEG data caused, for example, by
    changes in the subject's signals or the recording device itself, such as electrode placement and impedance. 
    To address this problem in traditional BCIs, adaptive systems are proposed to dynamically adjust their behavior and parameters based on changes in the user's mental state, the environment, or the input data quality.
   
    In the last years, several works have proposed using a specific brain signal elicited upon errors to improve BCIs. The so-called error-related potentials (ErrPs) can be elicited under different circumstances and measured with EEG~(for a review, see \cite{xavier_fidencio_error-related_2022, chavarriaga_errare_2014, kumar_review_2019}). 
    In BCI research, usually seven different types of ErrPs are mentioned. Errors committed by the subject are called \textit{response ErrPs}~\citep{blankertz_single_2002, van_schie_modulation_2004}. \textit{Feedback ErrPs} are generated upon feedback about a choice made~\citep{miltner_event-related_1997, chavarriaga_errare_2014} and \textit{target ErrPs} can be generated by implementing unexpected changes in the task~\citep{diedrichsen_neural_2005}. However, in BCI paradigms it is more common to find the use of either \textit{interaction ErrPs}~\citep{ferrez_you_2005, ferrez_error-related_2008}, which are elicited when the interface wrongly interprets the user's input, or \textit{observation ErrPs}, which are generated while the subject only observes a system over which they have no control perform a wrong action. 
    \textit{Execution and outcome ErrPs} are also reported as the neural response to unexpected movements \citep{diedrichsen_neural_2005, spuler_error-related_2015} or undesired outcomes \citep{krigolson_electroencephalographic_2008, spuler_error-related_2015, kreilinger_single_2016}, respectively. For an extensive review on each ErrP and experimental protocols used to elicit them see~\cite{xavier_fidencio_error-related_2022}. 
    
    ErrPs have been combined with reinforcement learning (RL) in different studies to improve BCI performance~(for a review, see~\cite{xavier_fidencio_error-related_2022}). In the RL framework, an agent learns by trial-and-error while interacting with an environment~\cite{Sutton2018}. The agent performs an action in the environment and receives a scalar numerical reward. Its goal is to maximize the cumulative reward, called return, and learn an optimal policy, that is mapping from inputs to actions. While supervised learning relies on ground-truth data being provided as a learning signal, RL agents need only a reward signal that indicates the quality of a policy to drive the learning process, making the use of ErrPs a natural fit, as they only represent the existence of an error, not what the expected outcome, action or observation would have been.
    In this work, because the agent's actions do not directly affect its next inputs, the setting is specifically a contextual bandit problem, where the agent's inputs are not representing states, as in the full RL problem, but rather a context that is independent between timesteps.

    This study introduces a novel ErrP and RL-based BCI framework for the development of adaptive BCIs. The framework applies reinforcement learning to learn the user intention directly from brain signals obtained with non-invasive recordings and uses the neural signature of errors measured in the form of interaction ErrPs to drive the learning. 
    Our hypothesis is that the intrinsic interactive nature of the RL framework is particularly suitable for the development of such systems, inspired by the work from~\cite{kim_intrinsic_2017}. Moreover, as ErrP are generated during human-system interaction upon BCI errors, it does not increase the mental load of the subject and directly constitutes a real-time feedback source for the RL agent. 

    To further validate the proposed framework, in this study we also introduce a novel hybrid BCI paradigm using motor imagery and ErrPs. We propose a relatively fast-paced game to improve subjects' motivation and engagement. The hypothesis is that the gamified version of the commonly used cursor control task can increase subjects' motivation and interest in using the BCI, increasing overall performance~\citep{skola_progressive_2019, atilla_gamification_2024}. 
    Moreover, the increased game speed better aligns with real-time decision-making scenarios requiring faster reactions from participants. 
    
    The rest of this work is structured as follows: Section~\ref{sec:related_work} reviews studies that have used ErrP for adaptive BCIs. We present the proposed ErrP-RL-based BCI framework and describe the datasets used in this study in Section~\ref{sec:methods}. Section \ref{sec:results} presents our results. Finally, we conclude this work with a brief discussion and overview in Section~\ref{sec:discussion}.

\section{Related Work}\label{sec:related_work}
    Adaptive BCIs using ErrPs have previously been proposed.
    \cite{llera_use_2011} introduced an adaptive logistic regression based on interaction ErrPs. The weights of the classifier were modified based on the ErrP classification results. The approach was validated using both simulated and MEG data for a two-class MI paradigm, showing significant performance improvements compared to the baseline static classifier. \cite{schiatti_effect_2019} later applied the same approach to MI data recorded with EEG. 
    \cite{mousavi_improving_2017} introduced a new strategy by directly combining the ErrP frequency-domain information and the MI-related modulations to improve the classification of MI trials. They used common spatial patterns (CSP) for feature extraction and linear discriminant analysis (LDA) for classification of ErrPs and MI, combining the results with a logistic regression and observed significant improvements in performance with the proposed framework. This approach was further validated in an online follow-up study~\citep{mousavi_hybrid_2020}. 
    
    The ErrP information have often been used to validate the output of the BCI classifier. 
    An online BCI-speller based on code-modulated visual evoked potentials (c-VEP) and ErrP was validated in \cite{spuler_online_2012}. In this study, c-VEP trials were classified using a support vector machine (SVM) and a spatial filter (cannonical correlation analysis). The ErrP information was used to label trials for the training dataset.
    \cite{artusi_performance_2011} considered the classification of movement-related cortical potentials (MRCPs) into different motor tasks (e.g., slow vs. fast arm flexion). They also used the ErrP information to label trials before adding them to a the training dataset for an SVM classifier.
    This approach was also used recently by \cite{tao_enhancement_2023} in a two-class MI task, using regularized common spatial patterns (R-CSP) for feature extraction and the combination of Fisher's discriminant analysis (FDA) and SVM for classification. Lastly, for the classification of MI data using k-NN, \cite{haotian_online_2023} also used the trials labeled based on the ErrP to create a dataset and, after applying cross-validation to evaluate MI classification improvement, they expanded the training dataset with the new trials. 
    \cite{chiang_closed-loop_2021} used a similar approach to show the benefits of including the ErrP feedback information in the the adaptation of a convolutional neural network (CNN) for the classification of steady state visually evoked potentials (SSVEPs) and \cite{wang_toward_2024} for MI classification. However, as these studies used three- and four-class problems, respectively, the training dataset only included trials that did not elicit ErrPs. 
    All these studies reported improved performance when using the ErrP-based adaptation.

\section{Materials and Methods}\label{sec:methods}
\begin{figure}[!ht]
            \centering
            \includegraphics[width=1\linewidth]{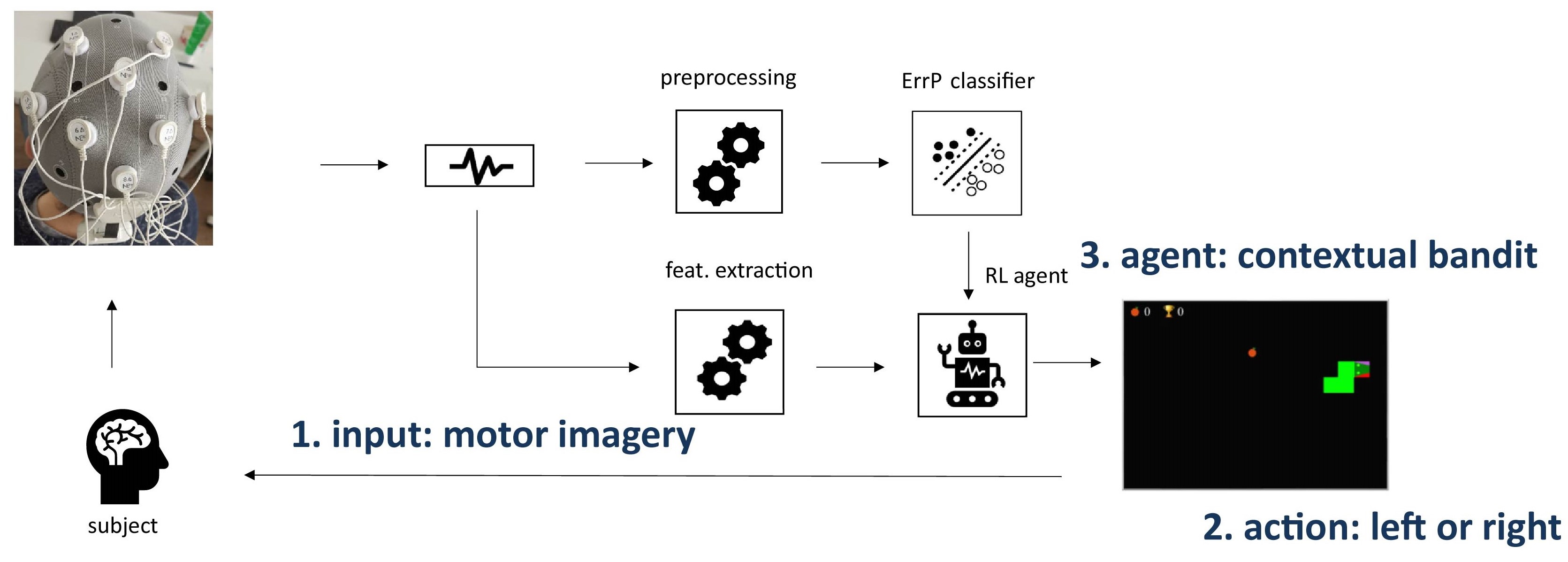}
            \caption{Overview of the BCI framework using ErrP and reinforcement learning. We consider non-invasive BCIs using EEG for neural signal acquisition. As proof-of-concept, we include BCIs based on motor imagery paradigms. The ErrP information is used as reward for the RL agent (3), which learns the mapping between motor imagery input features (1) and corresponding action (2) while subject plays the modified snake game.}
            \label{fig_MI_ErrP_RL_framework}
        \end{figure}

    We introduce a novel BCI framework using ErrP and reinforcement learning. 
    Figure~\ref{fig_MI_ErrP_RL_framework} shows the framework overview. Our hypothesis is that the intrinsic interactive nature of reinforcement learning agents is well-suited for the development of real-time adaptive BCIs. Moreover, ErrPs represent an intrinsic feedback source with no extra mental load given to the subject, as they are implicitly generated, even upon external error occurrences. Therefore, incorporating ErrP in the reinforcement learning framework as reward is very straightforward. The setup is validated offline with a motor imagery paradigm for BCIs.
    We used a well-known open-source dataset for a two-class motor imagery protocol. Additionally, we propose a new BCI task designed to combine the MI and ErrP paradigms in a gamified setup and use part of the data we collected to also validate the proposed RL-based framework. Details on all these components are described in the following sections.

    \subsection{Open-source dataset: BCI Competition IV dataset 2b}
        In this study, we used the open-source dataset 2b from the BCI Competition IV~\citep{leeb_braincomputer_2007}. This dataset is widely used as a benchmark for the classification of motor imagery. A detailed description of the experimental protocol can be found in the competition review~\citep{tangermann_review_2012}.
        
        The dataset includes EEG and electrooculogram (EOG) data from nine subjects recorded in five sessions. In the competition, the first three sessions were intended to be used for training and the last two for evaluation when validating proposed methods. However, different splits are commonly used in studies utilizing this dataset (for examples, see~\cite{ali_enhancing_2022}).
        In summary, the experimental task was a cue-based screening paradigm. 
        In the first two sessions, an arrow shown on the screen for $1.25$~s indicated the MI task that the subjects should perform (either left or right hand, with the movement freely chosen by each subject). Subjects had to imagine the corresponding hand movement for $4$~s. Afterwards, a break of at least $1.5$~s, followed by a randomized time of up to $1$~s, was included~\citep{tangermann_review_2012}. 
        The last three sessions included smiley feedback. Subjects were instructed to move this smiley towards the left or right side according to the cue and keep the MI for as long as possible. A break and a random interval were also included at the end of these trials. 
        EEG was recorded using three electrodes (C3, Cz, C4) with a sampling rate of 250~Hz. EOG was recorded using three monopolar electrodes. 
        Supplementary Table S1 reports the number of trials recorded for each subject. 

    \subsection{In-house dataset}
\begin{figure}[!ht]
                \centering
                \includegraphics[width=\textwidth]{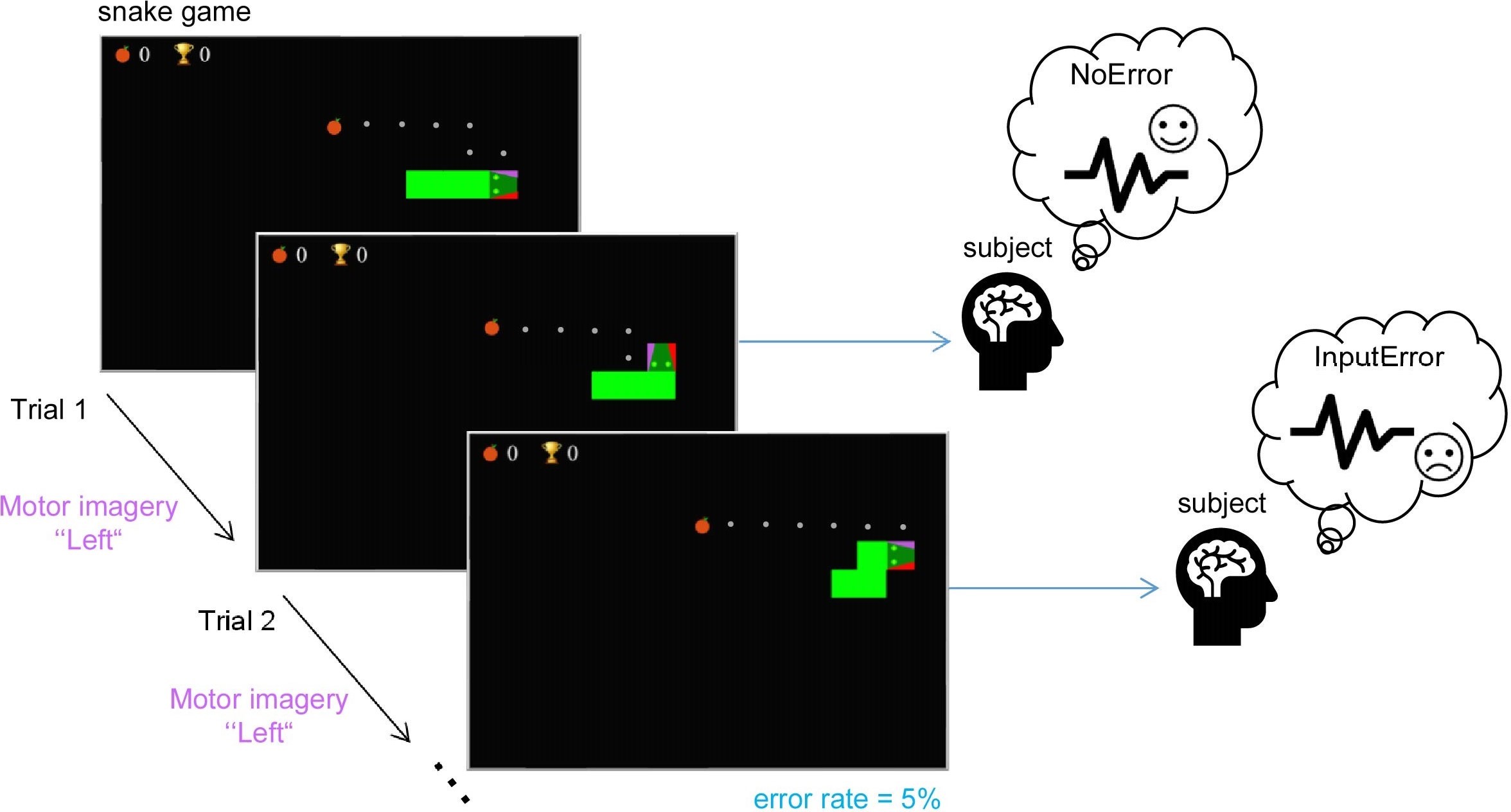}
                \caption{The experimental task: the subjects played the game by imagining a hand movement to control the snake and avoid collision with itself while following the given path (the displayed dots) to collect as many points as possible. In each trial in which the subject's input command was expected, with a probability of~$5$\%, the snake moved in the wrong direction (as depicted in Trial~$2$) to keep subjects motivated, but also to elicit ErrPs.}
                \label{fig_snake_MI_V0A}
            \end{figure}

        We implemented a new experimental protocol to validate the detection of motor imagery and ErrP-related neural signals.
        We developed a modified version of the snake game~\footnote{game used as baseline: https://www.geeksforgeeks.org/snake-game-in-python-using-pygame-module/} (see Figure~\ref{fig_snake_MI_V0A}) to (1) propose an interactive game design controlled via motor imagery to keep subjects motivated and focused; (2) demonstrate the feasibility of detecting MI-modulations with a fast-paced task; (3) demonstrate the feasibility of detecting interaction ErrPs in response to misinterpreted commands by the BCI at a low artificial error rate and simultaneously with the fast-paced MI control; (4) provide the basis for the development of an adaptive MI-based BCI using ErrP and reinforcement learning. 
        
        The study involving human participants was reviewed and approved by the Ethics Committee of Medical Faculty of the Ruhr University Bochum. The participants provided their written informed consent to participation.

        \subsubsection{Experimental protocol}

            All subjects were instructed both verbally and with written instructions to imagine the movement of the left or right hand to interact with the game. Participants chose freely whether to imagine an open-hand gesture or squeezing a ball. The game included a given path from the snake to the fruit that subjects were instructed to follow. This ensured the ground-truth label for the MI trials (left or right hand). Note that, while subjects believed they were actively controlling the snake, we did not decode the MI data online in this study. We artificially introduced error trials to keep subjects motivated. With a $5\%$ chance, the snake moved in the opposite direction as defined by the path. This also allowed us to in parallel demonstrate that interaction ErrPs can also be elicited with the proposed protocol (commonly used error rate in ErrP studies are 20-30\%, for a review see~\cite{xavier_fidencio_error-related_2022}). 
            We included a familiarization phase without artificial errors activation to let subjects get used to the game and avoid self-made errors (wrong hand movement imagined). Moreover, subjects were familiar with the snake game itself as they first participated in a recording session with the keyboard version of the game. In this first session, the subjects used two keys on the keyboard to interact with the game. The data of this session is not included in this study. 
    
            The recordings were performed in three phases. In a pilot study to validate the game, we recorded one subject and used the data to define the preprocessing pipeline. As the results with this subject were very promising, we extended to six other subjects. As the data analysis revealed significant modulations in the frequency domain as response to the motor imagination as expected only for two subjects, we decided to include the Movement Imagery Questionnaire-3 (MIQ-3)~\citep{williams_further_2012} to assess subject's ability to perform movement imagery before each recording. The study was then expanded to more subjects and we added a monetary compensation to attract participants. As none of the subjects had previous experience with motor imagery BCIs, we hypothesized that applying the questionnaire could help introducing the experimental protocol, further improving performance during the recordings. In this questionnaire, instructions are read to subjects to inform about the movement they first had to physically perform and then imagine (using either internal visual, external visual or kinesthetic imagery). After each mental task, they rated the ease/difficulty of performing the imagery on a 7-point Likert scale (1 - very hard to 7 - very easy to see/feel). Imagined movements included knee lift, jump, arm movement, and waist bending. 
            
            In total, thirty subjects participated in this study (thirteen male, age: $25.5 \pm 3.78$, one left-handed, all with normal or corrected-to normal vision). We excluded the data from eight subjects due to artifacted EEG data and one subject because of excessive movements during recordings. The data of the remaining twenty-one subjects (ten male, age: $26.0 \pm 4.21$, one left-handed) were analyzed. For each subject, between 600 and 900 trials were collected as recordings could be stopped at any time if subjects noticed they were losing focus. Subjects took a self-paced break between runs and were asked about continuing or ending the recording.
            
        \subsubsection{Data recording}
            EEG data was recorded with the Enobio wet EEG from Neuroelectrics at following positions: FC1, FC2, C3, Cz, C4, CP1, CP2, Pz. The sampling rate was set to $500$~Hz and CMS/DRL reference electrodes fixed behind the right ear. All recordings were performed in a quiet room and we turned off all electronic devices that were not required for the recording itself. The manufacturer's software uses a quality index (QI) instead of impedance control for the EEG channels. As recommended, recordings were only started when all channels showed a green or orange indicator. We implemented the use of a cotton swab soaked in skin-friendly disinfectant to remove hair between the electrode and scalp before gel application. Typically, all electrodes were green right after being filled with an appropriate amount of conductive gel.

        \subsubsection{Data analysis}
            EEG data was analyzed in MATLAB\textsuperscript{\textregistered} using the open source toolbox EEGLAB~\citep{delorme_eeglab_2004}. We performed a manual artifacted data rejection on continuous data, as recommended in the EEGLAB documentation to avoid spreading artifacts over good quality data. The data was then filtered with a Hamming windowed sinc FIR filter (0.5-100 Hz). As recommended in EEGLAB, low and high-pass filters were also applied separately. Additionally, a notch filter was applied to reduce power line noise. Lastly, the data was epoched and, if necessary, artifacted trials were removed. 
            Trials were extracted in the time interval [-1.0, 2.0]s around the snake's movement onset. We excluded epochs containing automatic forward movements of the snake. 
            Supplementary Table S2 reports the number of trials included for each subject.

    \subsection{Reinforcement learning agents}
        As stated in the introduction, the task faced by the agent in this setting, is not the full RL problem. This is because the agents input in timestep $t+1$ does not depend on the agents action in timestep $t$. Specifically, the subjects EEG data is unaffected by the agents classification of the data from the previous timestep.
        
        In this study, we have applied the contextual bandit algorithm LinUCB~\citep{li_contextual-bandit_2010} and its deep-learning counterpart, NeuralUCB~\citep{zhou_neural_2020}, to validate our framework. While we opted for these contextual bandit agents, it is important to note that the framework could accommodate other types of contextual bandit agents as well. The agents were implemented in python and are publicly available~\footnote{LinUCB: https://www.kaggle.com/code/phamvanvung/linucb/notebook | NeuralUCB: https://github.com/uclaml/NeuralUCB/ }. 
        
        The agents receive input data derived from human EEG signals. Their task is to learn the mapping between motor imagery-related features and intended actions. 
        This learning process receives direct feedback through interaction ErrPs, which can also be obtained from EEG data. A reward of 1.0 was assigned to an action when no ErrP was detected, and a reward of 0 was given otherwise. As in this study we only validate the framework offline, a perfect ErrPs classification was assumed using the true data labels which, in this case, are known beforehand. To simulate the online use of the proposed BCI framework, the motor imagery data was streamed trial-by-trial to the RL agents.
        
        We used optuna~\citep{akiba_optuna_2019} to optimize the hyperparameters of the agents for each subject individually. For LinUCB, $\alpha$ was searched over (0.01, 0.1, 1, 2, 4, 10) and for NeuralUCB we optimized the network hidden size ({16, 32, 64, 128, 256, 512}), $\nu$ ({0.1, 1, 10}), $\lambda$ ($10^{-i}, i = 1, 2 ,3 ,4$), and the learning rate ($2\times10^{-i}, 5\times10^{-i}, i = 1, 2, 3, 4$), as in~\cite{zhou_neural_2020}.
        
        The motor imagery features were extracted from the EEG data using continuous wavelet transform (CWT) following the methods used in several studies~\citep{ali_enhancing_2022, lee_application_2019}. With CWT, we can obtain a time-frequency representation of the EEG trials. The feature extraction was implemented in python using the library MNE-python~\citep{larson_2024_10999175} and the morlet wavelet was applied to a two-second epoch extracted from 0.5~s after cue-onset. Then we extracted both mu (6, 13)~Hz and beta (17, 30)~Hz bands power from the wavelet coefficients. This procedure results in a two-dimensional feature matrix (number of samples in frequency and time axes, respectively) with different dimensions for the frequency for the mu and beta bands. To achieve equal representation and avoid bias towards one frequency band, we resized the feature matrices to $15\times32$ using cubic interpolation as done in~\cite{ali_enhancing_2022} and \cite{lee_application_2019}. These features were extracted for all channels available (C3, Cz, C4) and combined into a single, flattened vector for the RL agent.

\section{Results}\label{sec:results}
    \subsection{BCI Competition IV dataset}
\begin{figure}
                \centering
                \includegraphics[width=1\linewidth]{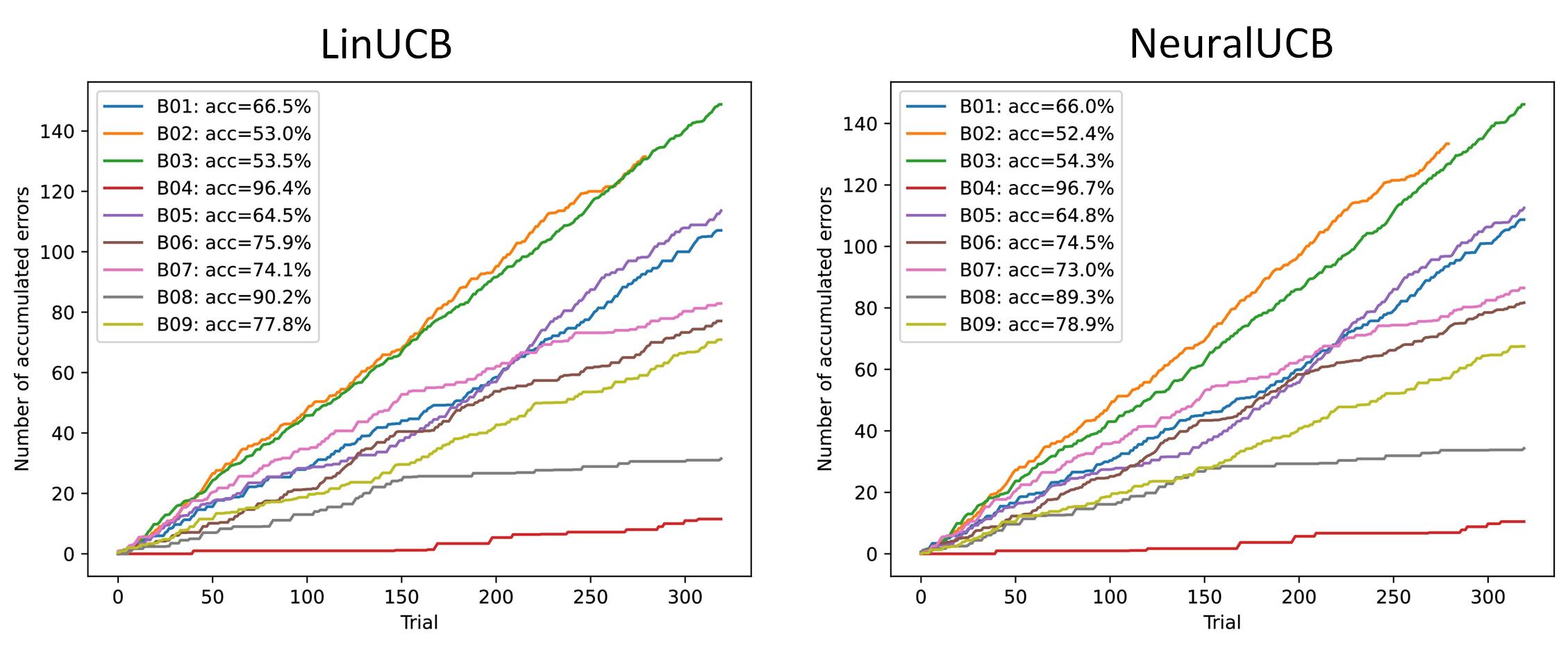}
                \caption{Performance of two contextual bandit agents (LinUCB and NeuralUCB) in the evaluation sessions for the open source dataset (n = 9). Results are averaged over different seeds (five and ten, respectively). The plots show the accumulated number of errors across all trials. The accuracy is calculated based on the final accumulated regret. Results show that both agents perform reasonably well for all except two subjects (B02, B03). There is no statistically significant difference in the performance of both agents (two-sided wilcoxon signed rank test, p = 0.91).}
                \label{fig_RL_BCI_IV_plots_both_agents_eval}
            \end{figure}

    The performance of the contextual bandit agents for all subjects in the open-source dataset is illustrated in Figure~\ref{fig_RL_BCI_IV_plots_both_agents_eval}
    We ran the agents for multiple random seeds and report the average results. As the LinCUB agent is CPU-bound, it was computationally expensive and we only executed five seeds. NeuralUCB benefits from GPU computation and we were able to use ten seeds. The results show that, for most subjects, both agents are able to learn the mapping of motor imagery input features into actions with reasonable accuracy. A two-sided wilcoxon signed-rank test for the accuracy shows no statistically significant difference between the two agents (two-sided p-value: 0.91). For two subjects (B02 and B03) both agents perform close to chance-level. This can be explained by the lack of features separability between the two motor imagery classes for these two subjects, which is directly reflected in the performance of supervised learning approaches in these datasets as well (see ~\cite{ali_enhancing_2022}). 
        
        We also evaluated the performance of the agents on the training data by splitting the trials in two (first- vs second-half of the trials) and testing whether its performance improved over time. One-sided wilcoxon signed-rank tests showed that, for both agents, the accumulated number of errors was significantly smaller in the second half of the training session compared to the first half (p=0.002, for both agents). This further indicates that the agents were able to learn while interactively receiving new motor imagery trials as input.

    \subsection{In-house dataset}
           
        \subsubsection{Neurophysiological analysis of the MI data}
\begin{figure}[!ht]
                    \centering
                    \includegraphics[width=\textwidth]{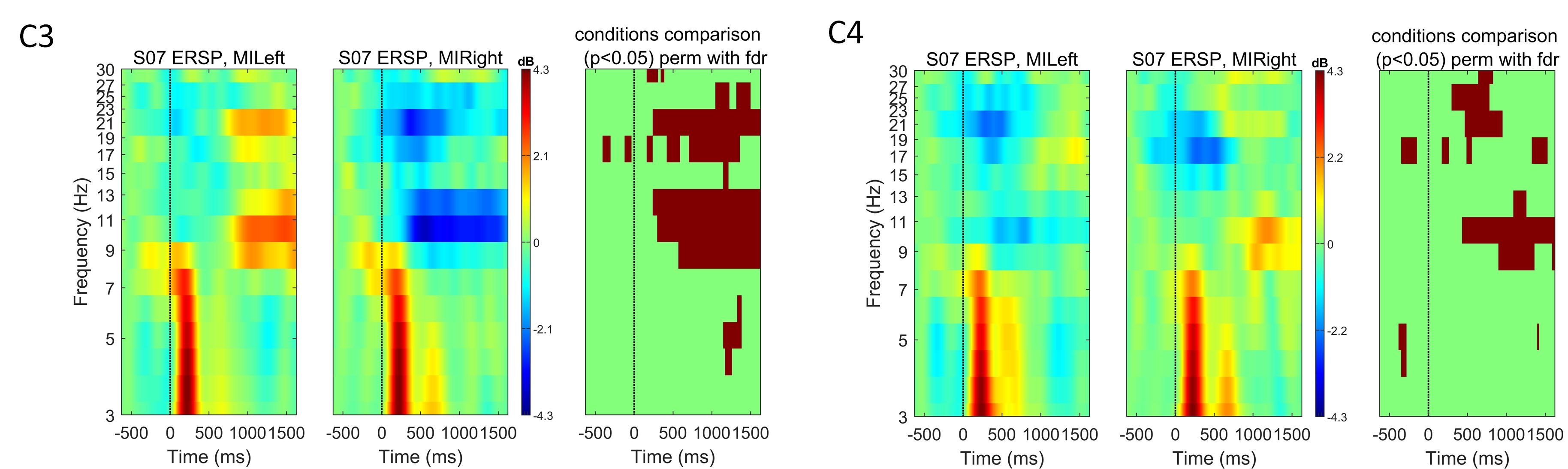}
                    \caption{Event-related spectral perturbation (ERSP) for one subject ($S07$) at channels C3 and C4 for both left and right hand motor imagery tasks. The color bars show the color and power spectral density in dB. For each channel, we used EEGLAB to compare the two experimental conditions (left vs right) and show in the right-most plot the permutation results (800 permutations, using false-discovery rate correction for multiple comparisons, significant p-values shown in red for $\alpha=0.05$). These plots highlight how motor-imagery related spectral modulations could be measured for this subject, with modulations mostly visible in the frequency ranges of (10-13)~Hz and (16-30)~Hz.}
                    \label{fig_S07_ersp_MI_V0A}
                \end{figure}

            The mean event-related changes in spectral power compared to the pre-stimulus baseline for the first subject used to validate the snake game with motor imagery control are depicted in Figure~\ref{fig_S07_ersp_MI_V0A}\footnote{Please note that this subject, assigned the ID S07, also participated in a previous ErrP-only study}.

\begin{figure}[!ht]
                    \centering
                    \includegraphics[width=\textwidth]{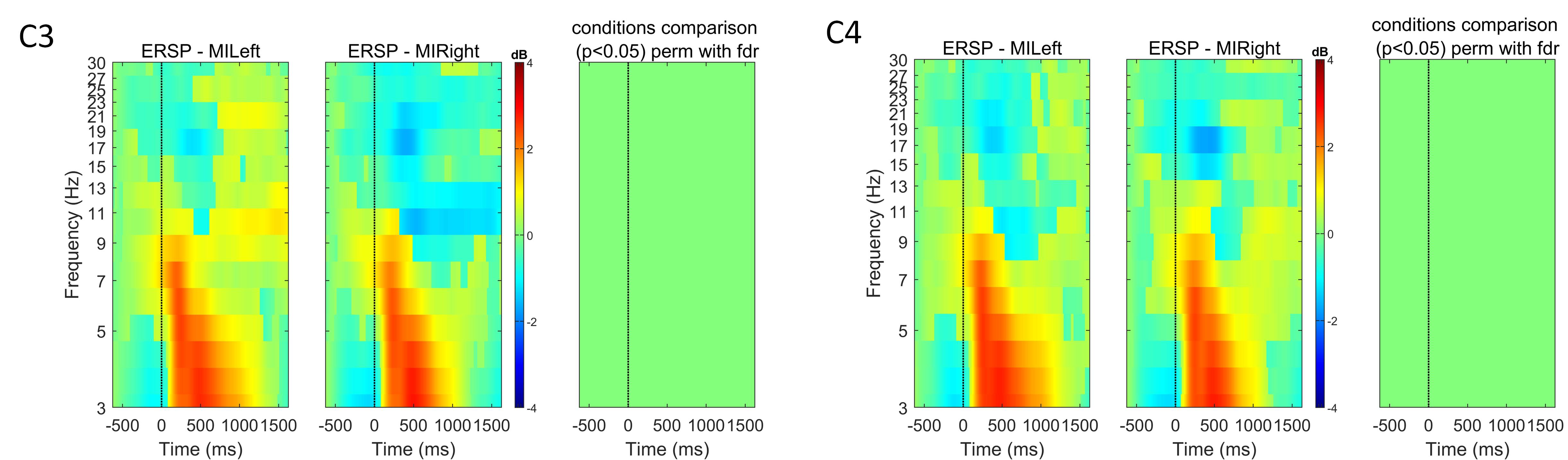}
                    \caption{Event-related spectral perturbation (ERSP) for all subjects (n~=~7) at channels C3 and C4 for both left and right hand motor imagery tasks. The color bars show the color and power spectral density in dB. For each channel, we used EEGLAB to compare the two experimental conditions (left vs right) and show in the right-most plot the permutation results (800 permutations, using false-discovery rate correction for multiple comparisons, for $\alpha=0.05$). These plots show that, some ERD is visible, the differences across experimental conditions are not significant when considering all subjects.}
                    \label{fig_ersp_pilot_MI_V0A_study}
                \end{figure}

\begin{figure}[!ht]
                    \centering
                    \includegraphics[width=\textwidth]{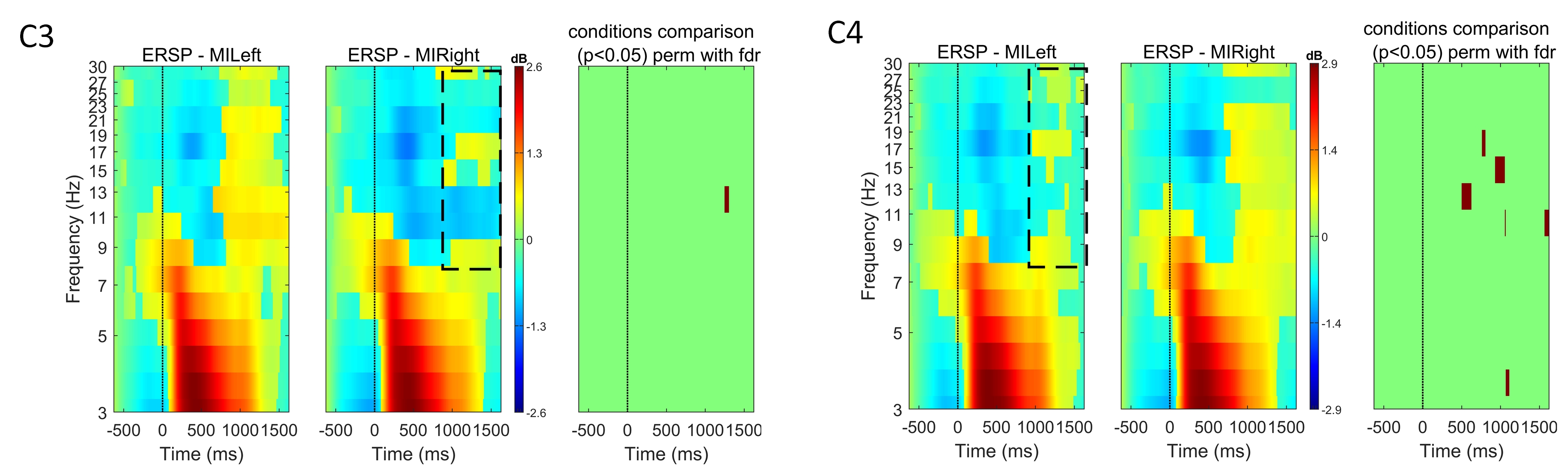}
                    \caption{Event-related spectral perturbation (ERSP) for all subjects (n~=~21) at channels C3 and C4 for both left and right hand motor imagery tasks. The color bars show the color and power spectral density in dB. For each channel, we used EEGLAB to compare the two experimental conditions (left vs right) and show in the right-most plot the permutation results (800 permutations, using false-discovery rate correction for multiple comparisons, for $\alpha=0.05$). These plots show that, some ERD is visible (see dotted areas for each condition in the contralateral hemisphere), the differences across experimental conditions are not significant when considering all subjects.}
                    \label{fig_ersp_all_MI_V0A_study}
                \end{figure}

            We used EEGLAB to generate the event-related spectral perturbation (ERSP) image. The following parameters were set: wavelet cycle parameter ([2, 0.1]), pre-stimulus baseline (-750, -500)~ms, and frequency range of (3, 30)~Hz. We look directly at channels C3 and C4, as MI-related modulations can be measured at electrodes located over the sensorimotor cortex~\citep{abiri_comprehensive_2019}. The ERSP images for this subject show the expected motor imagery-related modulation in the contralateral hemisphere. For example, for the right hand motor imagination (MIRight), an event-related desynchronization (ERD) is visible in the C3 electrode. This ERD is more pronounced in the frequency ranges within the expected mu (6, 13)~Hz and beta (17,30)~Hz bands~\citep{abiri_comprehensive_2019}. For each channel, we additionally show the statistical comparison of the ERSPs for the two experimental conditions, which highlight that the differences observed in mu and beta ranges are statistically significant (permutation test with 800 permutations and using false-discovery rate to correct for multiple comparison. In Figures~\ref{fig_S07_ersp_MI_V0A}, \ref{fig_ersp_pilot_MI_V0A_study} and \ref{fig_ersp_all_MI_V0A_study}, as well as Figures \ref{fig_S12_ERSP_diagrams_C3_C4} and \ref{fig_S29_ERSP_diagrams_C3_C4} in Appendix~\hyperref[sec:appendix]{A}, p-values below the significance level of $0.05$ are shown in red). 
              
            As described in Section~\ref{sec:methods}, we used the data recorded with this pilot subject to define the preprocessing pipeline and validate the feasibility of the experimental task for detecting motor imagery-related neural activity. Given the results shown in Figure~\ref{fig_S07_ersp_MI_V0A}, we extended the study to more subjects. 
            Figure~\ref{fig_ersp_pilot_MI_V0A_study} shows the mean event-related changes in spectral power for all subjects included in the first extended study (n~=~7), including $S07$. While ERSP plots show some contralateral ERD for the experimental conditions, the statistical comparison does not show significant differences. 
            
            Finally, Figure~\ref{fig_ersp_all_MI_V0A_study} shows the mean event-related changes in spectral power for all subjects included in the final extended study (n~=~21). Also in this case, while some ERD is visible, the differences across experimental conditions for all subjects are not significant as seen for the pilot subject ($S07$).
            
            The questionnaires to assess the motor imagery abilities were analyzed after the recordings and the scores are summarized in Supplementary Table S3. If we consider a threshold for the total motor imagery ability score at~$75\%$ of the maximum (score of $15.21$), only four subjects scored below this level. However, we found that a high score in the questionnaire did not imply significant ERSP modulations. Subject $S07$ did not reach the highest score and, still, for no other subject in this dataset such significant ERSP modulations were observed. Subject $S12$ obtained a low score, however, statistically significant differences were found in the ERSP modulations, especially during left hand motor imagination. On the other hand, subject $S29$ reached the highest score in the questionnaire but no significant ERSP modulations can be seen (see Figures~\ref{fig_S12_ERSP_diagrams_C3_C4} and~\ref{fig_S29_ERSP_diagrams_C3_C4} in Appendix~\hyperref[sec:appendix]{A}). Nonetheless, we believe that applying the questionnaire helped improving subjects' understanding of the concept of motor imagination and how it can be performed. However, one further aspect to consider is the increased overall experimental time when such questionnaire is applied. 
    
            Overall, the results obtained with the pilot subject $S07$ demonstrate the feasibility of using the proposed fast-paced motor imagery paradigm. On the other hand, it is intriguing to us that only one particular subject performed remarkably well. We extended the study first to seven and then to thirty subjects to validate the protocol with a broader audience with the expectation of having more subjects with such significant modulations. It is unclear to us how this particular subject differs from the others such that classification performance is so outstanding.
    
        \subsubsection{Neurophysiological analysis of the ErrPs}
            
\begin{figure}[!ht]
                \centering
                \includegraphics[width=\textwidth]{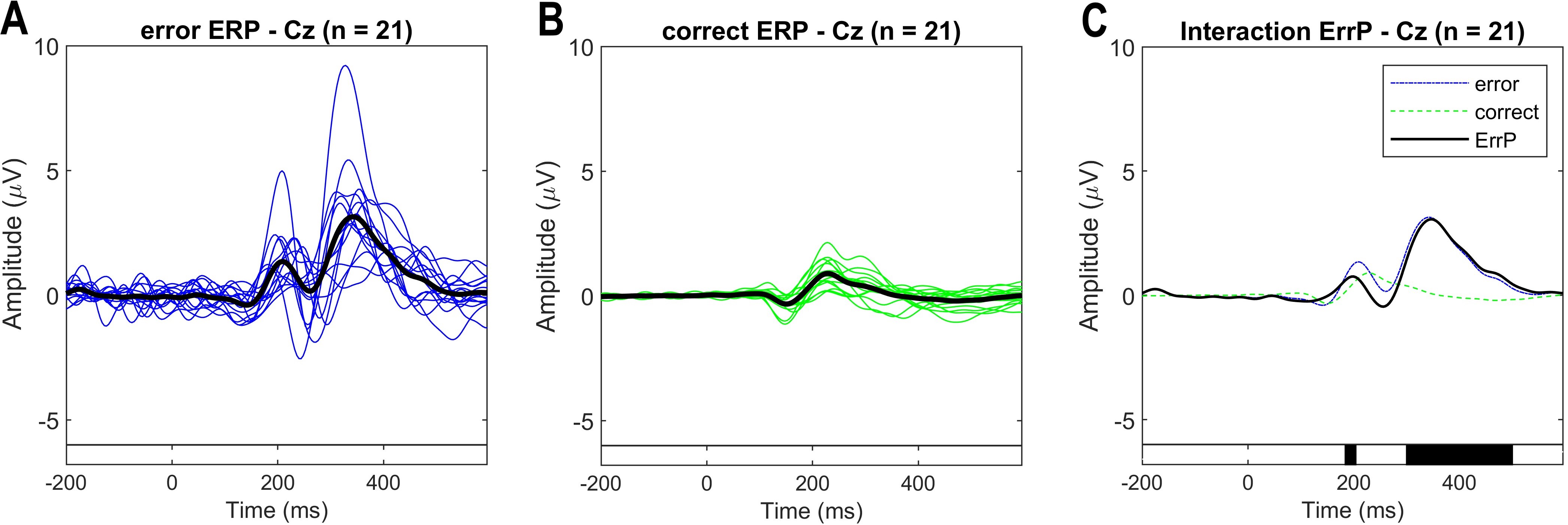}
                \caption{         
                (A) Event-related potentials (ERP) for the error condition for each subject (blue traces) and average over all (black trace). (B) ERP for the correct condition for each subject (green traces) and average over all (black trace). (C) ERPs at electrode Cz averaged over all subjects for each condition (error and correct). ErrPs are given as the difference grand average (error minus correct). On the bottom, the black background shows the time intervals with a significant difference between error and correct trials ($p<0.05$; corrected for multiple comparisons by false discovery rate (FDR) to avoid false positives). Results are consistent with related works, but a late negativity is not observed.}
                \label{fig_ErrP_Cz_MI_V0A_all_subjects_plots}
            \end{figure}

            As ErrPs are fronto-centrally located, higher amplitudes are expected at channels FCz and Cz~\citep{xavier_fidencio_error-related_2022}. The pre-stimulus interval [$-0.2, 0$]~s was used for baseline correction, and the ErrPs are calculated as the difference of error trials minus correct trials. Figure~\ref{fig_ErrP_Cz_MI_V0A_all_subjects_plots} shows the ErrP grand averages for the correct and error conditions measured at channel Cz.
            The measured ErrP displays a positive peak at $200$~ms, a negative peak at $252$~ms, and a positive peak at $348$~ms. Statistical comparison shows significant differences between the error and correct conditions. The observed waveform is consistent with existing literature on interaction ErrPs~\citep{ferrez_you_2005, ferrez_error-related_2008, ferrez_simultaneous_2008, ferrez_eeg-based_2009}. However, the expected negative peak between $430-550$~ms is not clearly visible with this experimental protocol. In our previous study when subjects interacted with the game via keypress this component was also visible in the ErrP (for details see~\cite{xavier_fidencio_generic_2024}).
    
        \subsubsection{Reinforcement learning results}
            Considering the results from the data analysis on the MI-related ERSP modulations for this study we only applied the agent to a subset of eight subjects. We selected subjects based on the presence of at least some significantly different modulations between left and right hand motor imagination, indicating the potential for sufficient class separability. Furthermore, considering the higher computation time for running the CPU-bound LinUCB agent, with the implementation used in this study, we only applied the NeuralUCB agent for this dataset, as results on the open source dataset were very similar between linear and neural UCB agents. 
            
\begin{figure}
                \centering
                \includegraphics[width=.5\linewidth]{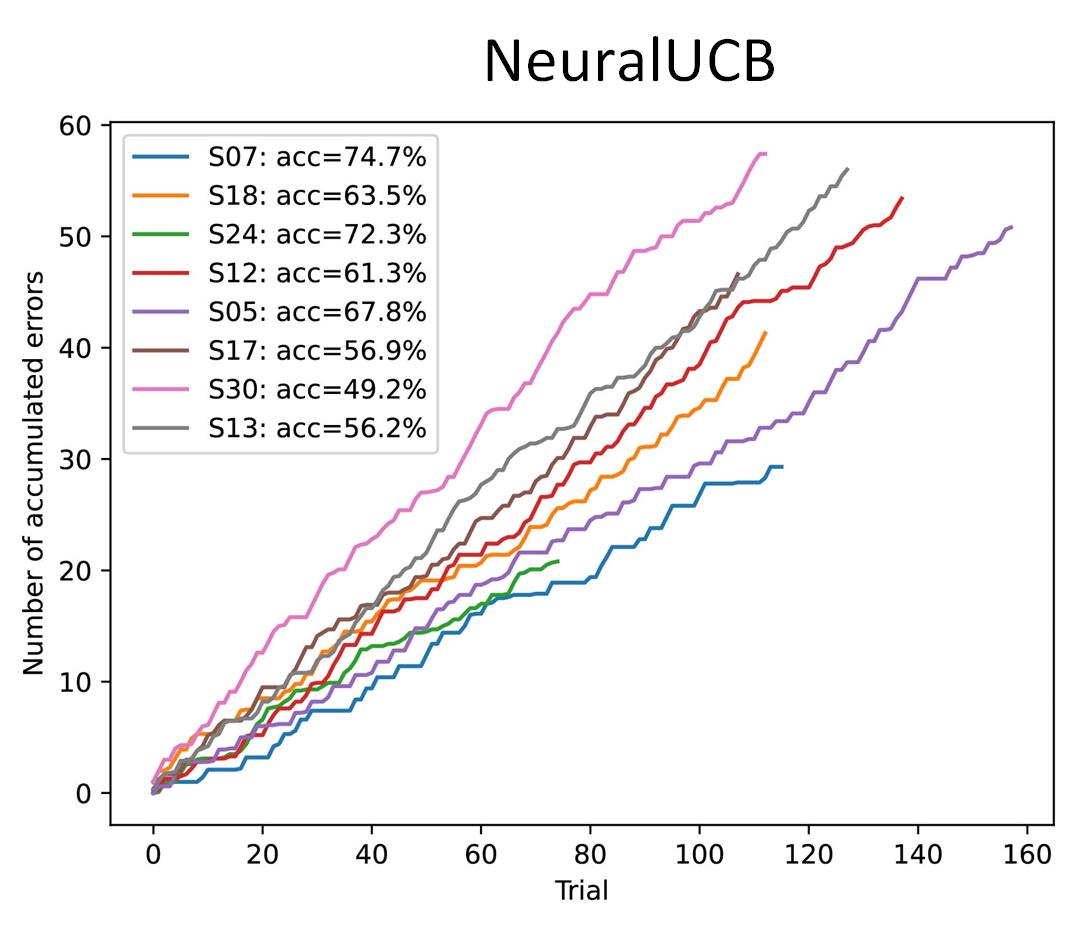}
                \caption{Performance of the NeuralUCB contextual bandit agent in the evaluation sessions for the in-house dataset (n~=~8). The plots show the accumulated number of errors across all trials, averaged over ten executions with different seeds. The accuracy is calculated based on the final accumulated regret. Results show that for most subjects the agent performance is not very high.}
                \label{fig_RL_MI_V0A_plot_neuralucb_eval}
            \end{figure}

            For each subject, the data was randomly shuffled and we used a simple train-test split to create the training and evaluation datasets ($80/20$). Figure~\ref{fig_RL_MI_V0A_plot_neuralucb_eval} shows the performance of the NeuralUCB agent in the evaluation datasets. 
            The pilot subject ($S07$) achieves the highest accuracy. This was expected and simply reflects the quality of the input features and the higher class separability.
            In general, the results obtained for subject S07 support our hypothesis that (1) the proposed experimental protocol can be used to elicit MI-related modulation and (2) a reinforcement learning agent can be used to learn the mapping between MI input features and intended action based on the feedback from the ErrP. On the other hand, the low class separability obtained for most subjects with our experimental protocol directly impacts the agent's performance, highlighting the need for high quality MI data in order to make the proposed ErrP-RL-based BCI framework feasible. 
            
\section{Discussion}\label{sec:discussion}
    This paper introduces a novel framework using error-related potentials (ErrPs) and reinforcement learning (RL) for the development of adaptive non-invasive brain-computer interfaces (BCIs). 
    This study explores the use of a contextual bandit agent to learn the mapping between motor imagery-related features and intended actions, demonstrating the feasibility and effectiveness of this approach in interpreting and responding to neural signals associated with motor imagery.
    The learning framework was applied to both an open-source and an in-house dataset we recorded using a new experimental protocol. The results indicate that the agents are able to learn from the time-frequency domain features extracted from EEG recordings with reasonable accuracy using the simulated perfect ErrP classification as reward. Moreover, the pilot study with the novel experimental task for using motor imagery (MI) and ErrPs in a fast-paced, interactive BCI suggest the feasibility of the introduced protocol. 
    
    Results with two contextual bandit algorithms (LinUCB and NeuralUCB) using the open-source BCI Competition IV dataset 2b for a two-class MI task show that both agents perform similarly and are able to learn the mappings between MI features extracted using the continuous wavelet transform, and the classes (left or right). 
    The results obtained with selected subjects recorded with the novel MI-ErrP experimental task confirms the feasibility of the proposed RL-based MI framework, complementing the results obtained with the open-source dataset. However, as in other domains of machine learning, reinforcement learning performance is highly dependent on the quality of features and, therefore, of the raw input data. The accuracies for individual subjects obtained in both datasets vary from close to chance level (e.g., for subjects B02 and B03 in the open-source dataset) to very high (e.g., for subject B04 in the open-source dataset). 
    Nevertheless, as the aim of this study was to validate the overall feasibility of the proposed learning framework, rather than obtaining optimal feature extraction performance, and performing MI can be a challenging task, we are confident that with higher quality MI features agents' performance can be further improved in the proposed framework.
    
    In this work, we used BCIs with a binary output. However, as the framework is based on reinforcement learning (RL), we believe the setup can be easily extended to non-binary tasks. This is, in fact, one of the main advantages of the proposed approach compared to the related works reviewed in Section~\ref{sec:related_work}. Most studies considered the binary case, where, upon ErrP detection, the true class label could be inferred as the opposite label (with some uncertainty due to the imperfect ErrP classification accuracy). In the few studies that considered more than two classes, only trials that did not elicit ErrPs were used, and trials for which the true label could not be inferred were discarded. In contrast, in an RL framework, such as the one proposed in this work, the agent can learn from every sample based on the reward received, regardless of the number of classes (actions) in the output.
    
    While RL shows potential for dealing with the non-stationarities in the EEG data due to its adaptability, its successful application in BCIs requires overcoming some challenges, such as noisy EEG data, limited training data, and the design of the reward function. In this work, the latter was addressed by including ErrPs as reward information, as it is intrinsically generated during interaction with BCIs. In the proposed framework, mistakes made by the BCI in the form of wrongly classified MI trials are expected to elicit an interaction ErrPs, which can be used to provide feedback for the agent in the form of reward (or penalty). On the other hand, as the learning results show, the agents performance is significantly reduced if the classes are less clearly separable. This is the case for subjects B02 and B03 in the open-source dataset, for whom other works also report reduced classification accuracies, and three out of eight subjects in the in-house collected dataset. Therefore, future work should further investigate different feature extraction methods to improve input data quality for the agents, and further validate the learning framework to establish performance boundaries and minimal requirements. 
        
    In this study we also introduce a novel experimental paradigm for the development of a hybrid BCI using motor imagery and ErrPs.  
    An interactive snake game was proposed to increase subject's motivation and mitigate issues like boredom and reduced attention that commonly happen in repetitive BCI tasks. 
    The increased game speed was defined based on previous studies and feedback from subjects, who were much less interested when playing the slower game. Moreover, a shorter interval better aligns with real-time decision-making scenarios that would require a faster reaction from subjects than commonly used slower-paced MI tasks. 
    Data collected with twenty-one subjects show the feasibility of eliciting ErrPs under a low error-rate of~$5\%$ while subjects perform MI in a fast-paced task. Data analysis of the MI data show the expected mu and beta band modulations, but significant differences across experimental conditions (left versus right hand motor imagination) are only visible for few subjects, with one subject performing particularly well. 
    
    While the used two-second step deviates from the commonly used timings (usually $3-5$~seconds), the proposed protocol provided insights into the feasibility of fast-paced MI tasks and the real-life deployment of such BCIs. On the other hand, such a short inter-trial interval might also not have been sufficient for most subjects to account for awareness of game state, cognitive preparation for the MI task, and execution of the MI task with robust neural activation, leading to the observed low class separability and consequently reduced decoding accuracy.
    Moreover, while subjects might be more motivated by playing, the continuous nature of the game might increase the cognitive load too much, leading to faster mental fatigue or inconsistent performance over time. 
    Furthermore, all recorded subjects had no previous experience with motor imagery. This can be learned and improved over several training sessions~\citep{tao_enhancement_2023}. Therefore, results indicate that future research should also validate the protocol with experienced subjects or include training sessions. Extending the experiment and analysis protocol this way should rule out some possible reasons for insufficient data quality, enabling systematic validation of the feasibility of the proposed experimental protocol. Another aspect that could be included is the evaluation of whether including a higher reaction time between feedback presentation and start of the motor imagination is required. Moreover, even though we already included breaks between experimental blocks, we would like to evaluate the quality of the motor imagination with shorter blocks (e.g., only~$20$ trials per block instead of~$120$).
    
    Another aspect from the proposed task to consider is that incorporating the pre-programmed paths in the game helped ensuring true-labels for the MI trials. This can also be particularly helpful for recording labeled data for training classifiers using supervised learning approaches, which is widely used in BCI development. On the other hand, if subjects doubt their influence on the game control, this can reduce their motivation and the quality of the MI data. 
    We also artificially introduced ErrPs in the task. In the envisioned online BCIs, these signals are generated because the BCI misinterpreted the subject's intention. In both cases, the interest in the BCI can also reduce if too many error trials are spotted.

    Lastly, in this study we analyzed the proposed framework offline and assumed a perfect ErrP classification. In reality, ErrP classification accuracy will most likely be lower. Our ongoing work also considers the systematic validation of the proposed framework considering different rates of ErrPs misclassification to understand the performance boundaries for a general ErrP-based RL framework for adaptive BCIs. Future work should validate the entire framework during online use, as this is the intended application of BCI systems.
    
    In summary, the development of non-invasive BCIs using EEG data usually requires the design of subject-specific classifiers to decode the neural modulations of interest for each specific paradigm. Not only must these classifiers be calibrated before use, but their performance might also degrade over time due to non-stationarities in the EEG signals. Some works have proposed different re-calibration strategies to update the classifiers and account for changes during long-term BCI use. In this work, we proposed and validated a new framework based on error-related potentials (ErrPs) and reinforcement learning for the development of adaptive BCIs. We hypothesize that RL methods have the potential of dealing with the non-stationarities of EEG signals and, by using the intrinsic ErrP generation as reward, they can constitute the fundamental block for the development on adaptive BCIs.

\section*{Conflict of Interest Statement}
The authors declare that the research was conducted in the absence of any commercial or financial relationships that could be construed as a potential conflict of interest.

\section*{Author Contributions}
AXF, FG and II defined the paper scope. AXF, CK and II conceived the experiment. AXF conducted the experiments, performed data analysis and wrote the first draft of the manuscript. All authors revised the manuscript.

\section*{Funding}
This work is supported by the Ministry of Economics, Innovation, Digitization and Energy of the State of North Rhine-Westphalia and the European Union, grants GE-2-2-023A (REXO) and IT-2-2-023 (VAFES).

\section*{Acknowledgments}
The authors would like to thank all subjects who participated in the study,  Lisa Gonsior and Marie Kranz for assisting with data collection, and Marie Schmidt and Susanne Dyck for the constructive discussions.

\section*{Data Availability Statement}
The dataset used for this study can be obtained from the corresponding author on a reasonable request.

\bibliographystyle{apalike}  
\bibliography{frontiers_bib}

\newpage
\section*{Appendix A}\label{sec:appendix}

    \begin{figure}[!ht]
        \centering
        \includegraphics[width=\textwidth]{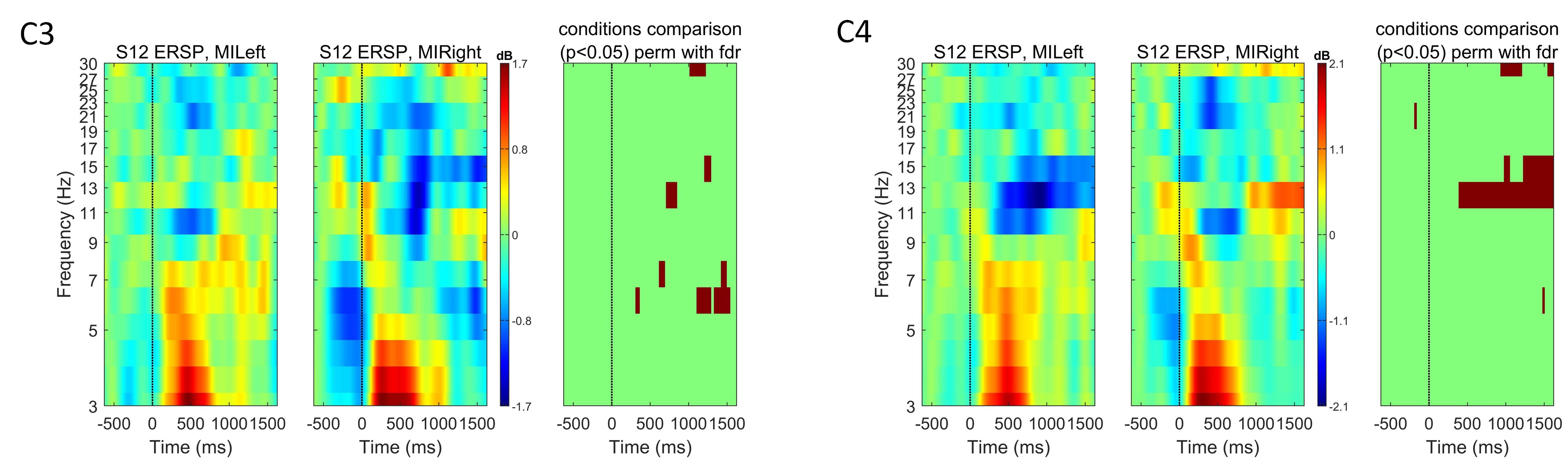}
        \caption{Event-related spectral perturbation (ERSP) for subject $S12$ at channels C3 and C4 for both left and right hand motor imagery tasks. The color bars show the color and power spectral density in dB. For each channel, we used EEGLAB to compare the two experimental conditions (left vs right) and show in the right-most plot the permutation results (800 permutations, using false-discovery rate correction for multiple comparisons, significant p-values shown in red for $\alpha=0.05$). These plots highlight how some motor-imagery related spectral modulations could be measured for this subject, even though they obtained the lowest score on the motor imagery ability compared to other subjects in the same study. }
        \label{fig_S12_ERSP_diagrams_C3_C4}
    \end{figure}

    \begin{figure}[!ht]
        \centering
        \includegraphics[width=\textwidth]{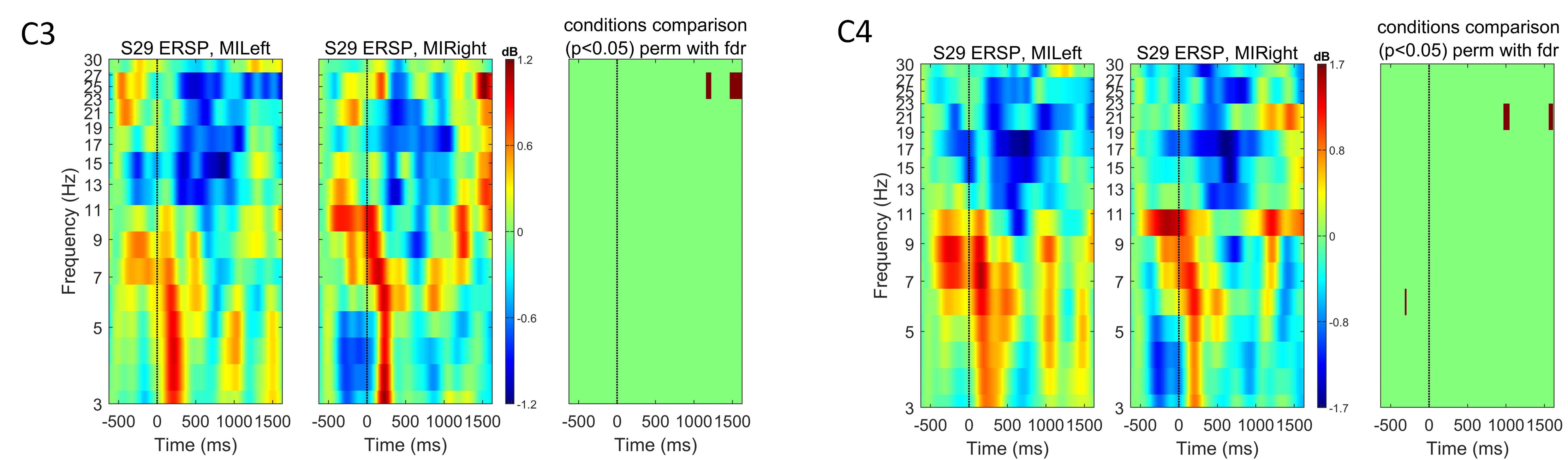}
        \caption{Event-related spectral perturbation (ERSP) for subject $S12$ at channels C3 and C4 for both left and right hand motor imagery tasks. The color bars show the color and power spectral density in dB. For each channel, we used EEGLAB to compare the two experimental conditions (left vs right) and show in the right-most plot the permutation results (800 permutations, using false-discovery rate correction for multiple comparisons, significant p-values shown in red for $\alpha=0.05$). These plots highlight how no significant motor-imagery related spectral modulations could be measured for this subject, even though they obtained the highest score on the motor imagery ability compared to other subjects in the same study. }
        \label{fig_S29_ERSP_diagrams_C3_C4}
    \end{figure}

\end{document}